\documentclass[]{interact}

\usepackage{epstopdf}
\usepackage{csquotes}              
\usepackage[style=apa, backend=biber, natbib=true]{biblatex}
\DeclareLanguageMapping{american}{american-apa}

\bibliography{references.bib}
\usepackage{amsmath}
\usepackage{threeparttable}

\usepackage{hyperref}
\usepackage{xurl}
\usepackage{mathtools}

\usepackage{amsthm}
\usepackage{color}
\usepackage{graphicx}
\usepackage{lmodern}
\usepackage{tablefootnote}
\usepackage{multirow}
\usepackage{subcaption}

\usepackage{tikz}
\usetikzlibrary{arrows.meta}
\usetikzlibrary{babel}
\usetikzlibrary{decorations.pathreplacing}
\usetikzlibrary{fit}
\usetikzlibrary{positioning}
\usetikzlibrary{calc}

\theoremstyle{plain}

\theoremstyle{definition}

\theoremstyle{remark}

\begin{document}

\articletype{ARTICLE}

\title{Digital-physical testbed for ship autonomy studies in the Marine Cybernetics Laboratory basin}

\author{
    \name{%
        Emir Cem Gezer\textsuperscript{*},
        Mael Korentin Ivan Moreau,
        Anders Sandneseng Høgden,
        Dong Trong Nguyen,
        Roger Skjetne,
        Asgeir Sørensen
        \thanks{*Corresponding author, Emir Cem Gezer. Email: emir.cem.gezer@ntnu.no} }
    \affil{Department of Marine Technology, Norwegian University of Science and Technology - NTNU - 7491 Trondheim, Norway}
}

\maketitle

\begin{abstract}
    The algorithms developed for Maritime Autonomous Surface Ships (MASS) are often challenging to test on actual vessels due to high operational costs and safety considerations.
    Simulations offer a cost-effective alternative and eliminate risks, but they may not accurately represent real-world dynamics for the given tasks.
    Utilizing small-scale model ships and robotic vessels in conjunction with a laboratory basin provides an accessible testing environment for the early stages of validation processes.
    However, designing and developing a model vessel for a single test can be costly and cumbersome, and researchers often lack access to such infrastructure.
    To address these challenges and enable streamlined testing, we have developed an in-house testbed that facilitates the development, testing, verification, and validation of MASS algorithms in a digital-physical laboratory.
    This infrastructure includes a set of small-scale model vessels, a simulation environment for each vessel, a comprehensive testbed environment, and a digital twin in Unity for visualization and remote control.
    With this, we aim to establish a design and verification pipeline that progresses from low-fidelity simulations, through potential-flow-based simulations of each vessel, to model-scale testing in the laboratory basin.
    Further advancement allows moving towards half-scale validation with R/V milliAmpere1 and full-scale validation with R/V Gunnerus.
    In this work, we present the testbed and its components and demonstrate its use for early-stage development of ship autonomy guidance, navigation, and control (GNC) algorithms.
    Model-scale hydrodynamic scaling, thruster characterization, and uncertainty propagation remain limitations for later-stage validation.
\end{abstract}


\section{Introduction}

Autonomous vessels have the potential to significantly influence the maritime sector by enhancing safety, operational efficiency, and economic viability~\citep{gu2021autonomous,munim2019autonomous}.
Recent advances in sensor technology, artificial intelligence, and marine cybernetics have spurred the development of prototype autonomous vessels of various sizes~\citep{schiaretti2017survey}.
Additionally, large-scale initiatives have emerged to demonstrate the real-world feasibility of uncrewed ships under diverse conditions~\citep{reddy2019zero}.
These efforts are largely driven by benefits such as increased safety (through reduced human error), crew shortage, lower operating costs, and reduced emissions, all of which hold significant promise for reshaping shipping operations and logistics~\citep{gu2021autonomous,psaraftis2023logistics}.

To bridge the gap between concept and full-scale deployment, researchers and industry leaders have developed or repurposed vessels of varying sizes as experimental platforms for autonomous navigation~\citep{schiaretti2017survey}.
The milliAmpere1 ferry~\citep{hinostroza2025milliampere1} is a half-scale prototype built to test autonomy functions in a real harbor environment, while its successor milliAmpere2~\citep{eide2025autonomous} is a full-scale ferry built for passenger transport.
Here, half-scale refers to the size and role of the vessel relative to an operational passenger ferry, not to geometric scaling of a parent design.
Both ferries have facilitated the development and validation of key autonomy functions such as sensor fusion, collision avoidance, and automated docking under real harbor conditions~\citep{hinostroza2025model,alsos2024lessons}.
Yara Birkeland~\citep{Skredderberget2025} is a fully electric container ship providing a full-scale test platform for uncrewed cargo operations.
Furthermore, the crewed research vessel R/V Gunnerus~\citep{marley2023four} is equipped with conventional ship systems and an extensive suite of sensors and data acquisition tools, and it provides a platform for advanced scientific studies.
At the commercial scale, companies like Maritime Robotics and SeaRobotics each offer a fleet of autonomous surface vessels (ASVs) of multiple sizes for robotics applications and hydrographic surveys, while Reach Subsea, Ocean Infinity, and Fugro each develop their fleets of autonomous vessels specifically designed for offshore operations.

Laboratory basins are widely used to test ship autonomy.
The reported setups share a common structure that includes a basin, an instrumented model vessel, and a control loop closed around it.
What differs is mostly what surrounds that loop.
Some operate with generated waves, wind, and current \citep{mahini2013experimental, dubey2020hardware}, or in ice \citep{wilkman2013forty}, while others run only in calm water \citep{he2024implementation, garofano2024obstacle}.
Some couple the basin to simulation, either through hardware-in-the-loop testing before wet runs \citep{dubey2020hardware} or by pairing the tanks with full-mission simulators \citep{he2024implementation}.
These testbeds are naturally tailored to the facility and the problems they were built to solve.

The technological maturity of the algorithms tested on the reported autonomous vessels varies, and it can be placed on the Technology Readiness Level (TRL) scale, which ranges from TRL~1 (basic concepts) to TRL~9 (proven, commercially deployed)~\citep{H2025}.
The half- and full-scale vessels described above typically operate within the mid- to high-end of this range.
Testing directly on them is generally more operationally extensive and costly, which limits their suitability for early, exploratory stages of algorithm development.
To balance realism with efficiency at early conceptual development, researchers often combine physical trials with software simulations \citep{xu2025review}.
Test basins with programmable wave makers and instrumentation infrastructure offer realistic sea conditions for evaluating control systems at lower TRLs.
However, access to such facilities is limited, and building model vessels can be costly and time-consuming.

\begin{figure}[h]
    \centering
    \resizebox{0.8\linewidth}{!}{\begin{tikzpicture}[
    font=\small,
    trlnode/.style={circle, draw=black!55, line width=0.6pt, minimum size=6.5mm,
            inner sep=0pt, font=\footnotesize\bfseries},
    call/.style={font=\footnotesize, align=center, text width=3.1cm,
            rounded corners=2pt, draw=black!25, fill=black!2, inner sep=2.5pt},
    arr/.style={-{Stealth[length=1.8mm]}, black!65, line width=0.6pt}
    ]
    \definecolor{labband}{RGB}{207,225,242}
    \definecolor{labhi}{RGB}{41,128,185}
    \definecolor{relband}{RGB}{253,212,158}
    \definecolor{opband}{RGB}{199,233,192}
    \definecolor{anaband}{RGB}{200,200,200}
    \definecolor{anabandpink}{RGB}{235,205,215}


    \fill[anabandpink!45] (0.6,-0.9) rectangle (3.0,0.5);
    \fill[labband!45]     (3.0,-0.9) rectangle (5.4,0.5);
    \fill[relband!45]     (5.4,-0.9) rectangle (7.8,0.5);
    \fill[opband!45]      (7.8,-0.9) rectangle (11.4,0.5);

    \draw[black!35, line width=0.6pt] (0.6,0) -- (11.4,0);

    \draw[-{Stealth[length=1.6mm]}, black!45, line width=0.5pt] (0.6,0.75) -- (11.4,0.75);
    \node[font=\footnotesize\itshape, text=black!60, anchor=west, fill=white, inner sep=1pt]
    at (0.6,0.75) {Technology Readiness Level (TRL)};

    \foreach \i/\c/\tc in {%
            1/anabandpink/black, 2/anabandpink/black,  3/labband/black,
            4/labband/black,
            5/relband/black, 6/relband/black,
            7/opband/black, 8/opband/black, 9/opband/black}{
            \node[trlnode, fill=\c, text=\tc] (s\i) at (1.2*\i,0) {\i};
        }

    \node[font=\footnotesize\itshape, text=black!60] at (1.8,-0.65) {Analytical};
    \node[font=\footnotesize\itshape, text=black!60] at (4.2,-0.65) {Laboratory};
    \node[font=\footnotesize\itshape, text=black!60] at (6.6,-0.65) {Relevant env.};
    \node[font=\footnotesize\itshape, text=black!60] at (9.6,-0.65) {Operational env.};

    \draw[decorate, decoration={brace, amplitude=8pt},
        black!55, line width=0.6pt]
    (1.2*1.6,1.0) -- (1.2*4.4,1.0)
    node[midway, yshift=0.8cm, call, text width=2.8cm] (csim)
    {Numerical models, \\\texttt{mcsimpy}, HIL sim.};

    \draw[decorate, decoration={brace, amplitude=8pt, mirror},
        black!55, line width=0.6pt]
    (1.2*2.7,-1.0) -- (1.2*5.3,-1.0)
    node[midway, yshift=-0.8cm, call, text width=3.1cm] (cgun)
    {Model-scale testing,\\MC-Lab, Cyberships};

    \draw[decorate, decoration={brace, amplitude=8pt},
        black!55, line width=0.6pt]
    (1.2*4.5,1.0) -- (1.2*7.0,1.0)
    node[midway, yshift=0.8cm, call, text width=3.1cm] (cmil)
    {Half-scale, Harbor,\\milliAmpere1};

    \draw[decorate, decoration={brace, amplitude=8pt, mirror},
        black!55, line width=0.6pt]
    (1.2*7-0.5,-1.0) -- (1.2*8,-1.0)
    node[midway, yshift=-0.8cm, call, text width=4.3cm] (cgun2)
    {Full-scale, Harbor, Open water, R/V Gunnerus, milliAmpere2};
\end{tikzpicture}}
    \caption{
        Technology Readiness Levels (TRLs) and their correlation with maritime autonomy research platforms and NTNU ocean technology laboratories.
        The TRL bands indicate how maturity evolves for the autonomy algorithms under test in each environment.
    }
    \label{fig:trl_levels}
\end{figure}
In response to this need, we have developed a digital-physical testbed that supports the design, verification, and validation of algorithms for Maritime Autonomous Surface Ships (MASS) with a growing fleet of model-scale Cybership (C/S) vessels.
Algorithms are first developed against TRLs 2-4 through rapid model-scale testing in various simulated environments, then evaluated against TRLs 3-5 through controlled basin testing, before moving on to the higher-level validation afforded by the half- and full-scale platforms (Figure~\ref{fig:trl_levels}).
Model-scale testing raises confidence in an algorithm but does not substitute for validation at full scale.

Moving along this path is not only a matter of scale.
Model-scale testing follows Froude scaling, which preserves the wave and rigid-body motion regime but leaves the Reynolds number far below full scale, distorting viscous damping and propulsor performance \citep{fossen2021handbook,liu2015literature}.
In practice, the control design and software architecture largely carry over to larger platforms, although the implementation must be adapted to each vessel, and the scale-dependent hydrodynamic and thruster models, as well as the controller gains, must be re-identified or retuned.
Model-scale results verify that an algorithm behaves as expected, not that a given tuning will perform identically at full scale.

The primary contribution is an open, modular software-hardware stack—featuring a hydrodynamic model library—that transforms the Marine Cybernetics Laboratory basin into a benchmark testbed for ship autonomy research.
The testbed combines simulation environments, hydrodynamic datasets, visualization tools, physical models, and a unified control interface intended for academic researchers and industry collaborators.
It includes simulation environments with varying levels of fidelity and a streamlined interface for early-stage development and testing, with all hydrodynamic data released publicly to support wider adoption.

\begin{figure}[h]
    \centering
    \includegraphics[width=0.9\textwidth, page=4]{figures/arch-crop.pdf}
    \caption{Overview of Marine Cybernetics Laboratory components.}
    \label{fig:testbed_overview}
\end{figure}

The paper is organized as follows.
Section \ref{sec:infrastructure} gives an overview of the testbed and details its physical infrastructure (Section \ref{sec:physical_infrastructure}), including the wave basin, model vessels, and hardware architecture of the C/S fleet.
Section \ref{sec:digial_infrastructure} describes the digital infrastructure, covering the simulation environments, the digital twin, and the control and estimation software, along with the data made available to the community.
Section \ref{sec:example_experiments} presents example experiments that showcase the testbed both for simulation and physical testing.
Finally, Section \ref{sec:conclusion} concludes the paper and outlines future work.

\section{Infrastructure} \label{sec:infrastructure}

\subsection{Overview}

The multi-layered infrastructure integrates digital simulations with a physical testing environment to create a progressive workflow for developing and validating MASS algorithms.
At the core of this setup is the C/S fleet (see Table~\ref{tab:date_of_vessels}), originally developed for education and research projects at NTNU.
To complement these models, we have developed corresponding simulation environments for each vessel.
These environments cover increasing levels of fidelity, from simple models to more detailed simulations with realistic wave loads, and support early-stage algorithm development and debugging.
We employ a digital twin environment in Unity that allows us to visualize and analyze vessel behaviors and study implications of remote operation \citep{halvorsen2024combining}.
These components form our design, verification, test, and validation pipeline, where algorithms are first studied in desktop simulations before being transferred to the marine laboratory for hardware-in-the-loop (HIL) simulations and physical testing on the model vessels.
\begin{table}[!h]
    \centering
    \begin{threeparttable}

        \begin{tabular}{l|l|c|l}
            \textbf{Vessel}             & \textbf{Appears on}                     & \textbf{Scale$^\dagger$} & \textbf{Status}   \\
            \hline
            Cybership I                 & \citet{pettersen1998underactuated}      & 1:70                     & End of life       \\
            Cybership II                & \citet{skjetne2005maneuvering}          & 1:70                     & Being retrofitted \\
            Cybership III               & \citet{berntsen2009ensuring}            & 1:30                     & End of life       \\
            C/S Enterprise I            & \citet{skaatun2011development}          & 1:50                     & Being retrofitted \\
            C/S Saucer                  & \citet{idland2015marine}                & 1:1                      & Operational       \\
            C/S Inocean Cat I Drillship & \citet{bjorno2016thruster}              & 1:90                     & Operational       \\
            C/S Voyager                 & \citet{elvenes2023development}\tnote{*} & 1:32                     & Operational       \\
            R/V milliAmpere1            & \citet{brekke2022milliampere}           & 1:1                      & Operational       \\
            R/V milliAmpere2            & \citet{eide2025autonomous}              & 1:1                      & Operational       \\
            R/V Gunnerus                & \citet{marley2023four}                  & 1:1                      & Operational       \\
        \end{tabular}
        \begin{tablenotes}
            \small
            \item[*]C/S Jonny has been renamed to C/S Voyager.
            \item[$\dagger$] Scale is the geometric scale relative to a parent full-scale design; 1:1 denotes a vessel that is not scaled from a parent design.
        \end{tablenotes}
    \end{threeparttable}
    \caption{Current status of the C/S fleet and research vessels with relevant references.}
    \label{tab:date_of_vessels}
    \vspace{-1em}
\end{table}

\subsection{Physical infrastructure}
\label{sec:physical_infrastructure}

The MC-Lab provides a well-instrumented environment for testing model-scale marine vessels in realistic wave conditions.
Its primary facility is a wave basin measuring $32\mathrm{m}$ from the wave maker to the beach and 6.4m wide, with a maximum depth of $1.5\mathrm{m}$.
Conductive and ultrasonic probes can be placed throughout the basin to monitor wave characteristics.
Position and attitude are measured by a Qualisys motion capture system with Oqus infrared cameras mounted along the basin, tracking passive reflective markers fixed to each vessel and streaming pose estimates to the onboard controllers in real time.

By 2032, the MC-Lab will be succeeded by a new model basin for academic research, called the Hydro-Cybernetics Laboratory in the House of Archimedes as part of the new Norwegian Ocean Technology Centre \citep{havteksenteret2026}, which will inherit the testbed reported in this paper.

\subsubsection{Wave maker}
\label{sec:wavemaker}

\begin{table}[h]
    \centering
    \begin{tabular}{ll}
        \hline
        \textbf{Parameter}           & \textbf{Value}           \\
        \hline
        Flap width                   & $6.4\mathrm{m}$          \\
        Hinge height above floor     & $\approx 0.75\mathrm{m}$ \\
        Maximum flap speed           & $0.59\mathrm{m/s}$       \\
        Maximum flap acceleration    & $7.6\mathrm{m/s^2}$      \\
        Maximum flap amplitude       & $0.25\mathrm{m}$         \\
        Maximum (usable) wave period & $1.5\mathrm{s}$          \\
        \hline
    \end{tabular}
    \caption{Operating limits of the MC-Lab wave maker.}
    \label{tab:wavemaker}
    \vspace{-1em}
\end{table}

Waves are generated by a single flap-type wave maker driven by a Bosch-Rexroth servo motor.
The flap spans the full basin width of $6.4\mathrm{m}$, is actuated from a single mid-span drive, and is bottom-hinged approximately $0.75\mathrm{m}$ above the basin floor.
It is capable of generating waves up to $0.25\mathrm{m}$ in amplitude at controlled frequencies, and a wave-damping beach is located at the far end that minimizes reflections for cleaner experiments.
A LabVIEW interface produces either regular waves or irregular sea states from operator-set parameters or replays user-generated drive files.
The principal operating limits of the wave maker are listed in Table~\ref{tab:wavemaker}, where the maximum amplitude refers to the flap stroke and the maximum period is the largest value for which wave quality remains acceptable.

\subsubsection{Model vessels}
\label{sec:model_vessels}

The vessels in the operational subset of the C/S fleet—C/S Enterprise I, C/S Saucer, C/S Inocean Cat I Drillship, and C/S Voyager—are the primary physical platforms for experimentation in the basin.
These vessels span a range of hull forms and scales, and their principal particulars are summarized in Table~\ref{tab:vessel_dimensions}.

\begin{table}[h]
    \centering
    \begin{tabular}{l|c|c|c|c}
        \textbf{Vessel}             & $m$ [kg] & $L$ [m] & $B$ [m] & $T$ [m] \\
        \hline
        C/S Enterprise~I            & 14.1     & 1.105   & 0.248   & 0.07    \\
        C/S Saucer                  & 3.8      & 0.396   & 0.396   & 0.03    \\
        C/S Inocean Cat~I Drillship & 92.7     & 2.444   & 0.423   & 0.110   \\
        C/S Voyager                 & 13.7     & 0.99    & 0.312   & 0.098   \\
    \end{tabular}
    \caption{Principal particulars of the operational C/S model vessels.}
    \label{tab:vessel_dimensions}
    \vspace{-1em}
\end{table}

\subsubsection{Hardware}
\label{subsubsec:hardware}

All vessels in the C/S fleet share the same hardware architecture designed for modularity, scalability, and reproducibility.
The primary difference between the vessels is the number and types of actuators and sensors.
The software stack accompanying the hardware allows all vessels to operate with minimal configuration differences.
As a result, new vessels can be added to the testbed environment with minimal re-engineering, with the existing architecture used as a template for both hardware and software.

\begin{figure}[h]
    \centering
    \includegraphics[width=0.70\textwidth, page=2]{figures/arch-crop.pdf}
    \caption{Shared hardware architecture across the C/S fleet. This architecture is designed for modularity, scalability, and reproducibility, allowing new vessels to be added to the testbed with minimal re-engineering.}
    \label{fig:vessel_hardware}
\end{figure}

At the core of each vessel is a Raspberry Pi single-board computer, which serves as the main onboard processor.
This choice offers a compact, capable, and well-documented platform for running control algorithms, handling sensor input, and managing actuator commands in real time.
Communication between onboard components is kept streamlined and maintainable by using standard hardware protocols, ensuring compatibility with a wide range of off-the-shelf sensors and devices; see Figure \ref{fig:vessel_hardware}.

\section{Digital infrastructure}
\label{sec:digial_infrastructure}

\subsection{Simulation environment}
\label{subsec:simulation_environment}

To support algorithm development and conceptual testing, our infrastructure includes three complementary simulation environments, each tailored for different stages of the development and verification process.
These simulation environments serve distinct purposes but are all aligned within our design and verification pipeline.
Because they use the same interface as the physical vessels, the control stack is agnostic to which of them is active.
A given test campaign selects one plant, and the same code runs unchanged against a simulator or against a model vessel in the basin.
The resulting data flow is shown in Figure~\ref{fig:dataflow}, where the selectable plant is represented as a switch rather than a fixed path, since the coupling between the digital and physical layers is configured per experiment.

\begin{figure}[h]
    \centering
    \resizebox{0.9\linewidth}{!}{\usetikzlibrary{arrows.meta,fit,positioning,calc}

\definecolor{labband}{RGB}{207,225,242}
\definecolor{labhi}{RGB}{41,128,185}
\definecolor{relband}{RGB}{253,212,158}
\definecolor{opband}{RGB}{199,233,192}
\definecolor{anaband}{RGB}{200,200,200}
\definecolor{physband}{RGB}{235,205,215}
\colorlet{line}{black!55}

\begin{tikzpicture}[
        font=\small,
        blk/.style={draw, rounded corners=3pt, line width=0.5pt,
                        minimum width=30mm, minimum height=12mm, align=center, inner sep=2pt},
        data/.style={blk, fill=anaband,  draw=anaband!45!black,  text=black!72},
        sim/.style={blk,  fill=labband,  draw=labhi,             text=labhi!72!black},
        phys/.style={blk, fill=physband, draw=physband!45!black, text=black!72},
        ctlst/.style={blk, fill=opband,  draw=opband!45!black,   text=opband!22!black},
        twin/.style={blk, fill=relband,  draw=relband!45!black,  text=relband!22!black},
        grp/.style={draw=line, dashed, dash pattern=on 2pt off 2pt,
                        rounded corners=5pt, line width=0.5pt, inner sep=4pt},
        grplbl/.style={font=\scriptsize\itshape, text=black!60, inner sep=2pt,
                        anchor=south west},
        arlbl/.style={font=\scriptsize\itshape, text=black!60, inner sep=2pt,
                        rotate=90, anchor=south},
        ar/.style={-{Stealth[length=2mm]}, line width=0.5pt, draw=line},
        bi/.style={{Stealth[length=2mm]}-{Stealth[length=2mm]}, line width=0.5pt, draw=line},
        ln/.style={line width=0.5pt, draw=line},
        ]

        \newcommand{\sub}[1]{{\scriptsize #1}}

        \def\ra{2.9}   \def\rb{0.775} \def\rc{-0.775} \def\rd{-2.9}
        \def\px{0.0}   \def\hx{4.3}   \def\tx{-2.6}
        \def\ctlx{6.2} \def\dtx{11.0} \def\cy{-0.775}
        \def\swl{2.5}  \def\rail{2.75} \def\piv{3.3} \def\jx{2.0}
        \def\pa{-0.475} \def\pb{-0.775} \def\pc{-1.075}

        \node[sim]  (mcs)  at (\px,\ra) {mcsimpy\\\sub{Potential-flow model}};
        \node[sim]  (shoe) at (\px,\rb) {shoeboxpy\\\sub{Simplified dynamics}};
        \node[sim]  (stf)  at (\px,\rc) {Stonefish\\\sub{Sensor and 3D models}};
        \node[phys] (veh)  at (\px,\rd) {Physical vessel\\\sub{Basin, waves, mocap}};
        \node[data] (hyd)  at (\hx,\ra) {Hydrodynamic data\\\sub{WAMIT, VERES, 3D hull scans}};

        \node[grp, fit=(mcs)]       (thybox)   {};
        \node[grp, fit=(shoe)(stf)] (simbox)   {};
        \node[grp, fit=(veh)]       (basinbox) {};
        \node[grplbl] at (thybox.north west)   {Theoretical work};
        \node[grplbl] at (simbox.north west)   {Simulation};
        \node[grplbl] at (basinbox.north west) {Basin};

        \draw[ln] (mcs.west) -- (\tx,\ra) -- (\tx,\rd);
        \draw[ar] (\tx,0)   -- (simbox.west);
        \draw[ar] (\tx,\rd) -- (basinbox.west);
        \fill[line] (\tx,0) circle (1.7pt);
        \node[arlbl] at ([xshift=-2pt]\tx,1.55) {Translation};

        \node[blk, fill=white, draw=line, minimum width=10mm] (sw) at (3.0,\cy) {};

        \foreach \y in {\pb,\pc}{
                        \draw[ln] (\swl,\y) -- (\rail,\y);
                        \draw[ln, fill=white] (\rail,\y) circle (1.5pt);}
        \draw[ln] (\swl,\pa) -- (\rail,\pa);
        \draw[line width=1.2pt, draw=line] (\piv,\cy) -- (\rail,\pa);
        \draw[ln] (\piv,\cy) -- (sw.east);
        \fill[line] (\rail,\pa) circle (1.7pt);
        \fill[line] (\piv,\cy) circle (1.7pt);

        \node[ctlst, minimum width=34mm] (ctl) at (\ctlx,\cy)
        {Control stack\\\sub{Observer, control, allocation}};
        \node[twin] (dt) at (\dtx,\cy) {Digital twin\\\sub{Visualization, remote control}};

        \draw[ar] (hyd.west) -- (mcs.east);
        \draw[bi, rounded corners=2pt] (shoe.east) -- (\jx,\rb) -- (\jx,\pa) -- (\swl,\pa);
        \draw[bi] (stf.east)  -- (\swl,\pb);
        \draw[bi, rounded corners=2pt] (veh.east)  -- (\jx,\rd) -- (\jx,\pc) -- (\swl,\pc);
        \draw[bi] (sw.east) -- (ctl.west);
        \draw[bi] (ctl.east) -- (dt.west);

\end{tikzpicture}}
    \caption{
        Configurable data flow of the testbed. The control stack communicates with one plant at a time, which is chosen by the requirements of the planned experiment.
        The simulation environments and the physical vessel share the same interface, allowing the control stack to operate unchanged across all platforms.
    }

    \label{fig:dataflow}
\end{figure}

\subsubsection{Low-fidelity model}
\label{subsubsec:low_fidelity_model_simulation}

We employ a lightweight simulation environment that approximates vessel dynamics using simplified hydrodynamics following the definitions in \citep{fossen2021handbook}.
This environment is primarily used for rapid testing of software interfaces and high-level control logic, as it shares the same programming interface as the physical platform.

Each vessel is modeled as a uniform rectangular prism with dimensions $L$, $B$, and $T$, giving a mass $m = \rho L B T$.
The principal moments of inertia are
\begin{align}
    I_x = \tfrac{1}{12} m (B^2 + T^2),
    I_y = \tfrac{1}{12} m (L^2 + T^2),
    I_z = \tfrac{1}{12} m (L^2 + B^2).
    \label{eq:reduced_order_inertia}
\end{align}
These define the diagonal rigid-body mass-inertia matrix $\mathbf{M}_{\mathrm{RB}}$.
Hydrodynamic effects are approximated by diagonal added-mass terms $\mathbf{M}_{\mathrm{A}}$ and linear damping coefficients $\mathbf{D}$.
Both are obtained by scaling the corresponding rigid-body values.
Typical values include an added mass of 20\% in surge and approximately 100\% in sway and roll.
Thruster dynamics are neglected for efficiency.
The 6DOF nonlinear equations of motion are
\begin{align}
     & \dot{\eta} = \mathbf{J}(\eta)\nu,                             \\
     & (\mathbf{M}_{\mathrm{RB}} + \mathbf{M}_{\mathrm{A}})\dot{\nu}
    + \bigl(\mathbf{C}_{\mathrm{RB}}(\nu) + \mathbf{C}_{\mathrm{A}}(\nu_r)\bigr)\nu_r
    + \mathbf{D}\nu_r
    =
    \tau + \mathbf{g}(\eta),
    \label{eq:motion_eq}
\end{align}
where $\eta \coloneqq [x,y,z,\phi, \theta, \psi]^\top$ is the vessel pose, $\mathbf{C}_{\mathrm{RB}}$ and $\mathbf{C}_{\mathrm{A}}$ are the rigid-body and added-mass Coriolis matrices, $\tau$ is the generalized load vector, and $\mathbf{J}$ maps body velocities to the inertial frame.
Let $\nu \coloneqq [\nu_{lin}^\top, \nu_{rot}^\top]^\top$ where $\nu_{lin} \coloneqq [u,v,w]^\top$, $\nu_{rot} \coloneqq [p,q,r]^\top$, and let $S(\cdot)$ denote the skew-symmetric operator.
The rigid-body Coriolis and centripetal matrix is then
\begin{align}
    \mathbf{C}_{\mathrm{RB}}(\nu) & =
    \begin{bmatrix}
        \mathbf{0}      & -m S(\nu_{lin})         \\
        -m S(\nu_{lin}) & -S(\mathbf{I}\nu_{rot})
    \end{bmatrix},
\end{align}
with $\mathbf{I}\nu_{rot} = [I_x p,\; I_y q,\; I_z r]^\top$.
The added-mass contribution is
\begin{align}
    \mathbf{C}_{\mathrm{A}}(\nu) & =
    \begin{bmatrix}
        \mathbf{0}                               & -S(\mathbf{M}_{\mathrm{A,lin}}\nu_{lin}) \\
        -S(\mathbf{M}_{\mathrm{A,lin}}\nu_{lin}) & -S(\mathbf{M}_{\mathrm{A,rot}}\nu_{rot})
    \end{bmatrix},
\end{align}
and the added-mass and total mass matrices are
\begin{align}
    \mathbf{M}_{\mathrm{A}}  =
    \begin{bmatrix}
        \mathbf{M}_{\mathrm{A,lin}} & \mathbf{0}                  \\
        \mathbf{0}                  & \mathbf{M}_{\mathrm{A,rot}}
    \end{bmatrix},\quad
    \mathbf{M}               = \mathbf{M}_{\mathrm{RB}} + \mathbf{M}_{\mathrm{A}}.
\end{align}
For small roll and pitch angles, a diagonal restoring term $\mathbf{g}(\eta)$ provides linear restoring forces in heave, roll, and pitch.
Numerically, we integrate $\eta$ and $\nu$ using a fourth-order Runge--Kutta step at each timestep $\Delta t$.
Although this simplified model omits detailed thruster and environmental modeling, it offers sufficient fidelity for early-stage control system verification and greatly reduces computational overhead.

\subsubsection{Higher-fidelity potential-flow model}
\label{subsubsec:high_fidelity_simulation_model}

The potential-flow-based numerical model \emph{mcsimpy}, written in Python, approximates the dynamics of the vessels in the C/S fleet by combining rigid-body dynamics with frequency-dependent hydrodynamics and wave-excitation models.
It is built on the studies by \citet*{mo2023real,hygen2023deterministic} as well as the Marine Systems Simulator~\citep{perez2006overview}.
Here, hydrodynamic coefficients such as the added mass, radiation damping, and wave-excitation loads are derived from potential-flow analyses of the vessel hulls.
Zero-speed coefficients and response amplitude operators are computed with the boundary-element solver WAMIT~\citep{wamit}.
Because the model is based on potential-flow theory, it does not resolve viscous or strongly nonlinear effects.
Irregular sea states are generated using the JONSWAP and PM~\citep{pierson1964proposed} wave spectra.
\emph{mcsimpy} supports both 3 and 6 degrees of freedom (DOF) models.

This level of modeling deliberately targets sufficient fidelity required for guidance, navigation, and control (GNC) algorithm development and controller tuning, with or without wave, wind, and current disturbances.
At the same time, it is capable of running in real time with a significant number of wave components enabled \citep{hygen2023deterministic}.
Potential-flow-based models capture the dominant inertial, radiation, and wave-excitation effects governing vessel motion.
The model has been coded with emphasis on computational efficiency to support repeated simulation and evaluation for meta-learning and reinforcement learning studies \citep{roen2025meta, MScpdal26A}.
The publicly released dataset accompanying this work includes the 3D hull visualization, hydrodynamic coefficients, response amplitude operators, and wave-excitation load parameters.

\begin{figure}[!h]
    \centering
    \begin{subfigure}[t]{0.52\textwidth}
        \centering
        \includegraphics[height=5.1cm,keepaspectratio]{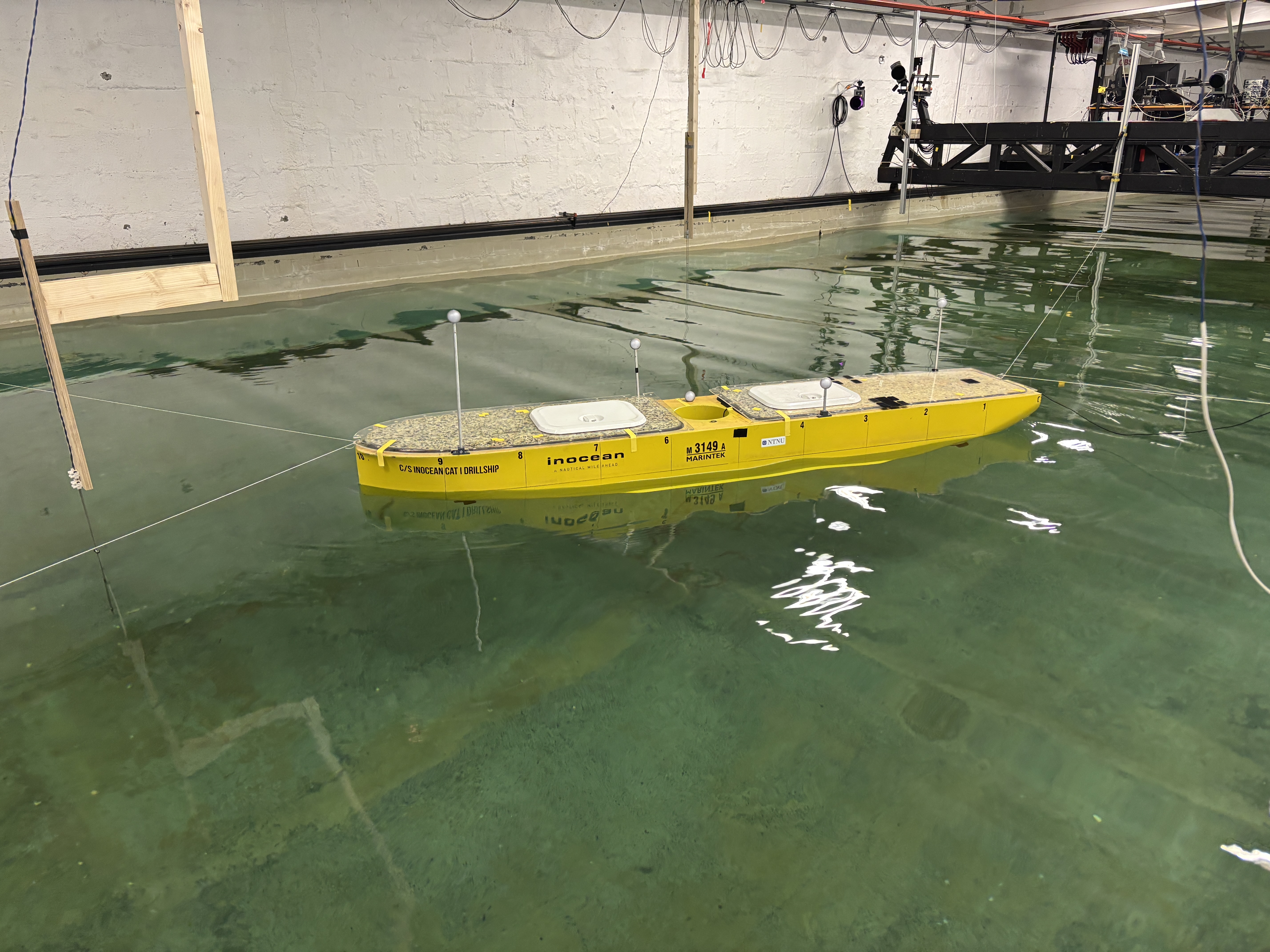}
        \caption{}
        \label{fig:mooring_setup}
    \end{subfigure}
    \hfill
    \begin{subfigure}[t]{0.47\textwidth}
        \centering
        \includegraphics[height=5.1cm,keepaspectratio]{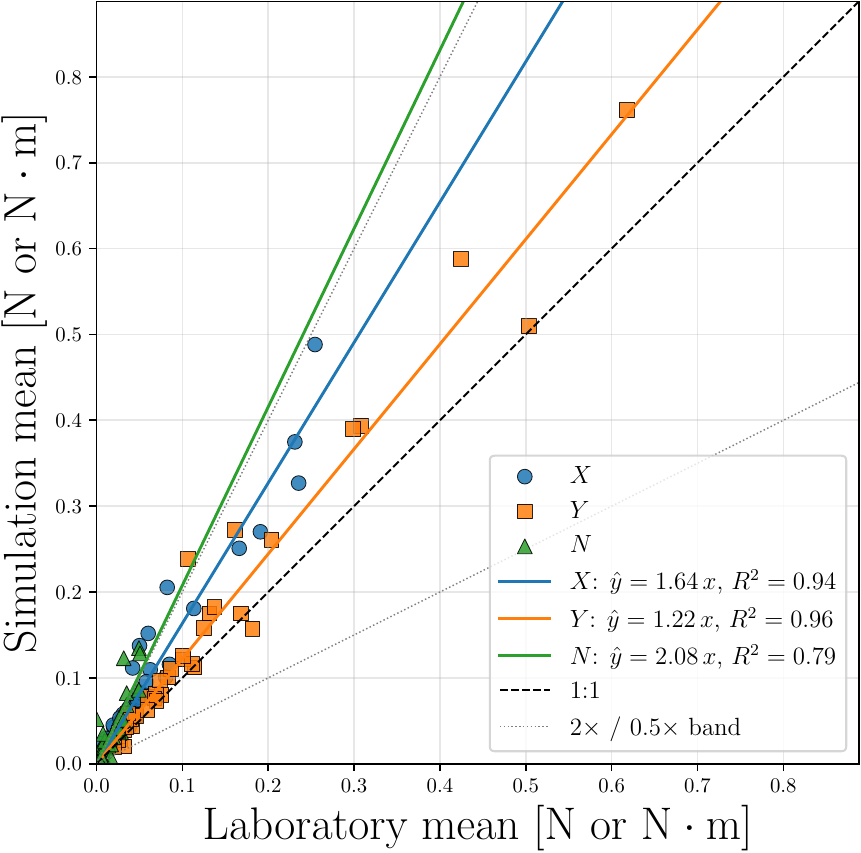}
        \caption{}
        \label{fig:mooring_results_parity_plot}
    \end{subfigure}
    \caption{
        Mooring test setup and validation.
        (\subref{fig:mooring_setup}) Physical mooring configuration in the MC-Lab basin.
        (\subref{fig:mooring_results_parity_plot}) Simulated versus measured mean surge (X), sway (Y), and yaw (N) load magnitudes across 44 laboratory tests.
        Per-DOF regression slopes show that the simulation systematically overpredicts the measured loads, and the agreement does not establish absolute hydrodynamic compliance.
    }
    \label{fig:mooring_test}
\end{figure}

Laboratory experiments have been conducted to compare the potential-flow model pipeline with the real system responses.
We used the C/S Inocean Cat I Drillship as a case study.
In these experiments, we placed four mooring lines on the vessel as shown in Figure \ref{fig:mooring_setup}, and ran 44 test cases with varying irregular sea states, both in simulation and in the physical basin, spanning significant wave heights $H_s = 0.5$-$5.5$~cm and peak periods $T_p = 0.71$-$1.61$s.

The results are summarized in Figure~\ref{fig:mooring_results_parity_plot}, which shows the mean surge, sway, and yaw loads measured in the mooring lines in the physical basin versus those predicted as mean wave loads in the simulation.
We calculated the mean absolute error (MAE) to be 0.038~N in surge, 0.029~N in sway, and 0.018~Nm in yaw, and the root-mean-square error (RMSE) to be 0.059~N, 0.049~N, and 0.029~Nm, respectively.
The simulation systematically overpredicts the physical loads, and this can be explained by discrepancies in the original WAMIT analysis that gave the hydrodynamic coefficients used by the simulation model: (i) the original model had a drilling turret that has been sealed off in the actual vessel, (ii) the analysis was done for a vessel LoA of 2.58~m, which is longer than the actual 2.444~m, and (iii) the draft and mass of the original vessel were significantly higher than those used in the basin tests.
Despite these discrepancies, the simulation captures the relative trend in loading across the sea states, most clearly in surge and sway.
This level of agreement supports the use of the potential-flow model for relative load trend prediction and early-stage control design and tuning, but it does not establish absolute hydrodynamic accuracy for other vessels or loading conditions.
Ongoing work targets updated hydrodynamic analyses to minimize the gap between the physical and digital models.

\subsubsection{3D simulation with sensor models}
\label{subsubsec:simplified_dynamics_with_sensor_models}
Stonefish~\citep{cieslak2019stonefish} is a high-performance and real-time capable robotics simulator written in C++.
Here, it serves mainly as a sandbox for image processing, and it also simulates other sensors such as inertial measurement units and sonars, with simplified contact and wave dynamics.
Stonefish is ideal for evaluating algorithms like localization, object detection, and platform interaction that require environmental variability and realistic sensor feedback.
It is integrated into the pipeline to bridge the gap between numerical simulation models and physical testing, allowing perception and control algorithms to be evaluated in a virtual environment.

\subsection{Digital Twin}
\label{subsec:digital_twin}

A digital twin environment has been developed for visualization and control of vessels in a virtual representation of the MC-Lab~\citep{premraj2023development}.
It consists of Unity-implemented 3D models of the MC-Lab and selected vessels from the C/S fleet, including the C/S Enterprise I and C/S Saucer.
The digital twin environment can be used in conjunction with the aforementioned simulation environments or with the physical vessels in the MC-Lab.
In this work, the term digital twin refers to a bidirectionally coupled visualization and remote-control environment rather than an automatically self-updating predictive model.
Vessel state is streamed from the simulators or the physical vessels into the virtual environment, while the operator sends manual commands, waypoints, and speed references back to the vessel through the same interface.
This setup enables the monitoring and demonstration of control algorithms in a more intuitive format to support testing in early development phases.

The digital twin system is paired with a remote control center, which allows operators to control the physical vessels in the MC-Lab from a separate location \citep{halvorsen2024combining}.
The vessel can be commanded manually using joysticks or Metaquest 2 headset and hand controls\footnote{Link to the video demonstration: \url{https://www.youtube.com/watch?v=fqEdp8VXAqE}}, as well as using these interfaces to provide guidance commands like waypoints and speed references for automatic control.
Vessel state information, such as position and orientation, is transmitted over the site-to-site network to provide remote monitoring capabilities.
While some limitations exist due to hardware constraints---particularly in the responsiveness of the monitoring---the system has been tested in both simulation and physical experiments and has been shown to function reliably under the available conditions.

\subsection{Software}
\label{subsec:software}

\subsubsection{Core modules}
\label{subsubsec:core_modules}

\begin{figure}[!th]
    \centering
    \includegraphics[width=0.7\textwidth, page=3]{figures/arch-crop.pdf}
    \caption{Control architecture of the testbed environment showing the closed-loop system: sensors feed an EKF observer that drives controllers, whose outputs route through thrust allocation to actuators. A multiplexer allows switching between control modules.
    }
    \label{fig:control_architecture}
\end{figure}

All the vessels in the C/S fleet operate on a shared software stack which is written in Python and C++ programming languages.
Robot Operating System 2  (ROS 2) serves as the middleware layer, enabling efficient inter-process communication and seamless integration with additional components and external programs ~\citep{macenski2022robot}.
This middleware facilitates modularity, allowing various control, sensing, and estimation modules to operate independently while maintaining synchronized communication across the system.

\subsubsection{Control and estimation modules}
\label{subsubsec:control_and_estimation_modules}

Each vessel runs a set of default software modules, allowing for immediate testing and validation of standard behaviors.
These include a thrust allocation module for distribution of forces to the thrusters, a proportional-integral (PI) feedback velocity tracking controller, a third-order reference filter integrated with a PID dynamic positioning controller, and finally an extended Kalman filter (EKF) from the \emph{robot localization} package \citep{MooreStouchKeneralizedEkf2014} as well as the availability of model-based observers with bias estimation and wave filtering.
Here, the EKF is used mainly to filter out the noise in the sensor measurements.

The C/S fleet shares a unified and modular control architecture that simplifies integration into the broader testbed infrastructure and provides a consistent foundation for algorithm development.
Built-in control and estimation modules offer baseline performance metrics that serve as reference implementations for testing and benchmarking new strategies.
Depending on the experimental need, these modules can be disabled or replaced with custom implementations, allowing researchers to test their own algorithms in a consistent environment.
To this end, we chose to keep the default modules as simple as possible, while still providing the necessary functionality for basic control and estimation tasks.
The overall control architecture is illustrated in Figure \ref{fig:control_architecture}.
Using this architecture, we can easily test and validate various control algorithms consistently.

\textbf{Thrust allocation}\\
We employ the maneuvering-based dynamic thrust allocation method~\citep{gezer2024maneuvering}, which formulates the thrust allocation task as a dynamic control problem.
It uses a control Lyapunov function (CLF) based filter design, respecting rate limitations, and a control barrier function (CBF) to handle the force saturations, while ensuring smooth and feasible thrust commands.

\textbf{Velocity tracking control}\\
A PI with reference feedforward (PI-RFF) velocity controller is made available to track a desired velocity.
For the body-frame low-speed velocity $\nu := [u,v,r]^{\top} \in \mathbb{R}^{3}$, the 3-DOF surge--sway--yaw plant with Coriolis and centrifugal terms neglected, is
\begin{equation}
    \mathbf M\dot{\nu} + \mathbf D\nu = \tau ,
    \label{eq:plant}
\end{equation}
where $\tau \in \mathbb{R}^3$ is the thrust load vector, $\mathbf M = \mathbf M^\top > 0$, and $\mathbf D = \mathbf D^\top> 0$.
Define the tracking error $\tilde\nu \coloneqq \nu-\nu_d$ for reference velocity $\nu_d(t)$, and the integral state $\dot\xi=\tilde\nu$.
The desired acceleration
\begin{equation}
    a_d \coloneqq \dot{\nu}_d - \mathbf{K}_p\tilde\nu - \mathbf{K}_i\xi ,
\end{equation}
with $\mathbf{K}_p,\mathbf{K}_i \in \mathbb{R}^{3\times3}$ PI gain matrices, yields the commanded load
\begin{equation}
    \tau_{\mathrm{cmd}} \coloneqq \mathbf M a_d + \mathbf D \nu_d,
    \label{eq:control_law}
\end{equation}
which combines inertial reference load feedforward $\mathbf M\dot\nu_d$, PI feedback, and feedforward compensation of damping using $\mathbf D\nu_d$.
Letting $z \coloneqq [\xi^\top, \tilde{\nu}^\top]^\top$, the closed-loop system becomes
\begin{align}
    \dot{z} = \mathbf{A}z + \mathbf{B}\mu, \quad \mu \coloneqq - \mathbf{K} z, \\
    \mathbf{A} \coloneqq \begin{bmatrix}
                             \mathbf{0}_{3\times3} & \mathbf{I}_{3}             \\
                             \mathbf{0}_{3\times3} & -\mathbf{M}^{-1}\mathbf{D}
                         \end{bmatrix}, \quad
    \mathbf{B} \coloneqq \begin{bmatrix}
                             \mathbf{0}_{3\times3} \\
                             \mathbf{I}_{3}
                         \end{bmatrix},
\end{align}
where the state feedback gain $\mathbf{K} \coloneqq [\mathbf{K}_i, \mathbf{K}_p]$ is designed so that $\mathbf{A} - \mathbf{B}\mathbf{K}$ is Hurwitz.
$\mathbf{K}_i$ and $\mathbf{K}_p$ can then be found using standard linear techniques such as pole placement or LQR synthesis.

\textbf{Dynamic Positioning controller}\\
We employ a mass-damper-spring-based reference model for the Dynamic Positioning (DP) controller.
The reference filter, in accordance with \citep{fossen2021handbook}, defines a state vector $x_d(t) \coloneqq [\eta^\top_d(t),\dot{\eta}^\top_d(t),\ddot{\eta}^\top_d(t)]^\top$, where $\eta_d \in \mathbb{R}^3$ is the desired pose in the world frame.
The filter evolves according to the continuous-time state-space model
\begin{align}
    \dot{x}_d(t) = \mathbf{A}_d x_d(t) + \mathbf{B}_d \eta_r,
\end{align}
where $\eta_r(t)$ is the reference steady-state pose input.
The filter matrices are defined as
\begin{align}
    \mathbf{A}_d
    \coloneqq
    \begin{bmatrix}
        \mathbf{0}_{3\times3} & \mathbf{I}_3                                                  & \mathbf{0}_{3\times3}                                         \\
        \mathbf{0}_{3\times3} & \mathbf{0}_{3\times3}                                         & \mathbf{I}_3                                                  \\
        -\mathbf{\Omega}^3    & -\bigl(2\mathbf{\Delta} + \mathbf{I}_3\bigr)\mathbf{\Omega}^2 & - \bigl(2 \mathbf{\Delta} + \mathbf{I}_3\bigr)\mathbf{\Omega}
    \end{bmatrix},
    \qquad
    \mathbf{B}_d
    \coloneqq
    \begin{bmatrix}
        \mathbf{0}_{3 \times 3} \\
        \mathbf{0}_{3 \times 3} \\
        \mathbf{\Omega}^3
    \end{bmatrix},
\end{align}
with $\mathbf{\Omega} \coloneqq \mathrm{diag}(\omega_1,\omega_2,\omega_3)$ controlling the filter bandwidth, and $\mathbf{\Delta} \coloneqq \mathrm{diag}(\delta_1,\delta_2,\delta_3)$ controlling the filter damping.
A forward-Euler step is used to update the filter state to provide desired pose, velocity, and acceleration outputs for the DP controller.

The default controller uses a PID control law to generate the commanded thrust loads for the vessel to track the reference trajectory.
Defining the pose error $\tilde{\eta} := \eta - \eta_d$ and the integral state
$\dot{\xi} \coloneq \tilde{\eta},$
the commanded thrust load $\tau_{cmd} \in \mathbb{R}^3$ is assigned as
\begin{align}
     & \tau_{cmd} \coloneqq
    - \mathbf{K}_p\mathbf{R}(\psi)^\top \tilde{\eta}
    - \mathbf{K}_d\mathbf{R}(\psi)^\top \dot{\tilde{\eta}}
    - \mathbf{K}_i \mathbf{R}(\psi)^\top \xi,
    \\
     & \mathbf{R}(\psi)
    \coloneqq
    \begin{bmatrix}
        \cos \psi & -\sin \psi & 0 \\
        \sin \psi & \cos \psi  & 0 \\
        0         & 0          & 1
    \end{bmatrix},
\end{align}
where $\mathbf{K}_p = \mathbf{K}_p^\top > 0$, $\mathbf{K}_i = \mathbf{K}_i^\top > 0$, and $\mathbf{K}_d = \mathbf{K}_d^\top > 0$ are the P, I, and D gain matrices, respectively.
The integral term $-\mathbf{K}_i \mathbf{R}(\psi)^\top \xi$ compensates for slowly varying or constant unmodeled disturbances (e.g., current or thruster bias) that the PD terms alone cannot correct for, at the cost of requiring standard anti-windup measures on $\xi$ during actuator saturation.

\subsection{Available data}
\label{subsec:available_data}

To allow for reproducibility and to support the development of new simulations, we have generated and publicly released comprehensive hydrodynamic datasets.
We performed stationkeeping hydrodynamic analyses using WAMIT~\citep{wamit}, a boundary-element software grounded in potential flow theory.
The submerged portions of each hull were meshed from 3D scans with some geometric simplifications.
To optimize computation, we simulated only half of each hull by exploiting ship symmetry.
From these runs, we obtained hydrostatic coefficients, linear hydrodynamic loads, second-order mean drift forces and moments, and response amplitude operators (RAOs) for the six rigid-body motions.
Linear hydrodynamic loads (added mass and damping) were derived from the radiation problem, while wave excitations were computed from the diffraction problem over headings ranging from $0^\circ$ to $180^\circ$.
The RAOs were based on vessel mass alone, excluding mooring or viscous effects.

\section{Examples of digital-physical test campaigns}

\subsection{4-corner DP test}
\label{subsec:four_corner_validation}

\label{sec:example_experiments}
\begin{figure}[h]
    \centering
    \begin{subfigure}[b]{0.48\textwidth}
        \centering
        \includegraphics[width=\textwidth]{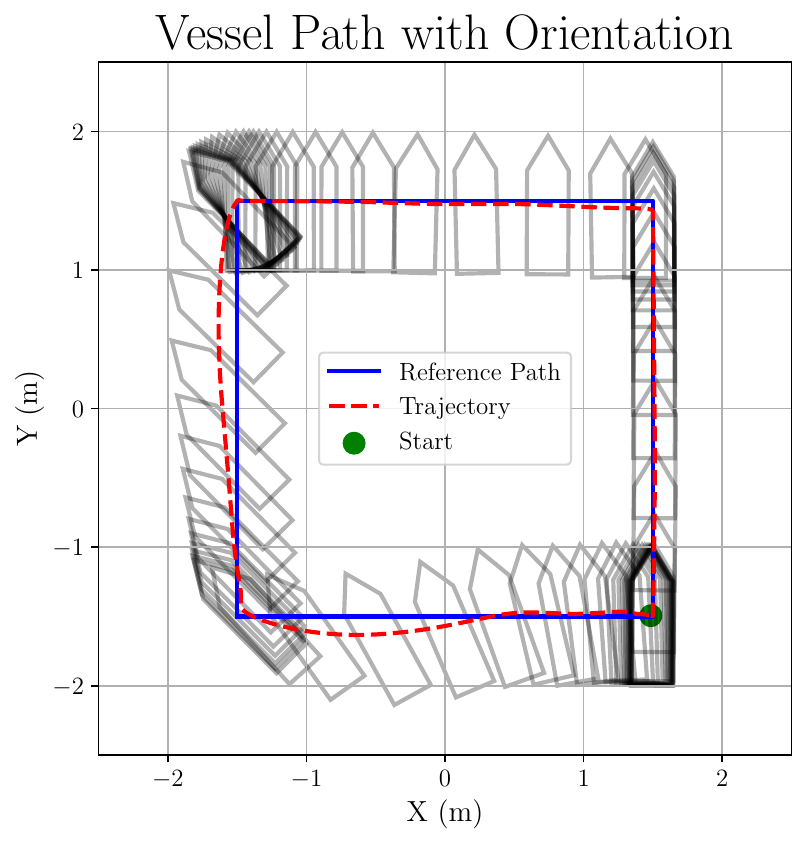}\\
        \includegraphics[width=\textwidth]{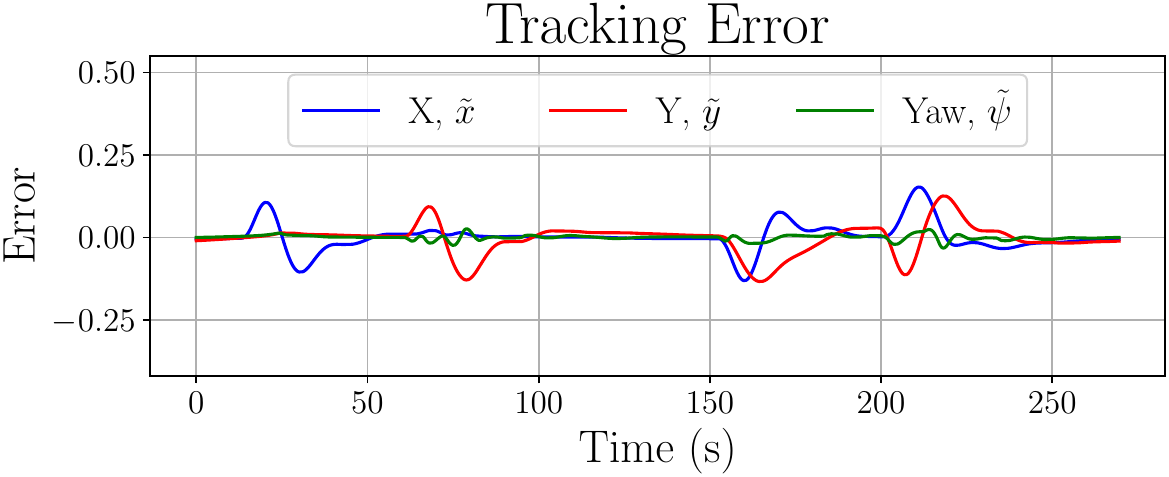}
        \caption{Reduced-order simulation}
        \label{fig:voyager_sim_subfig}
    \end{subfigure}
    \hfill
    \begin{subfigure}[b]{0.48\textwidth}
        \centering
        \includegraphics[width=\textwidth]{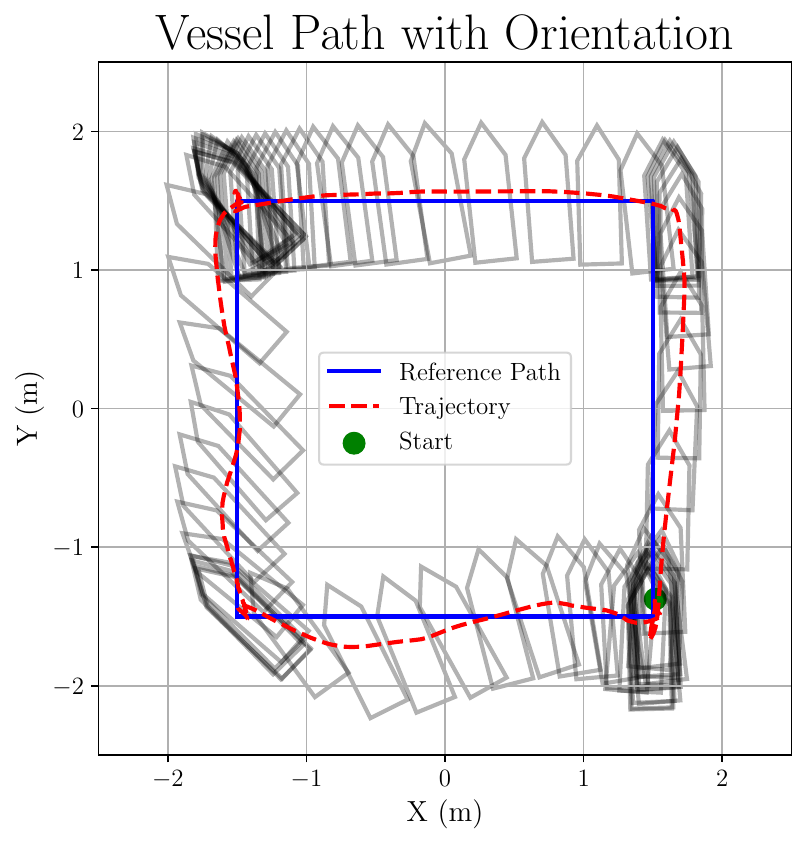}\\
        \includegraphics[width=\textwidth]{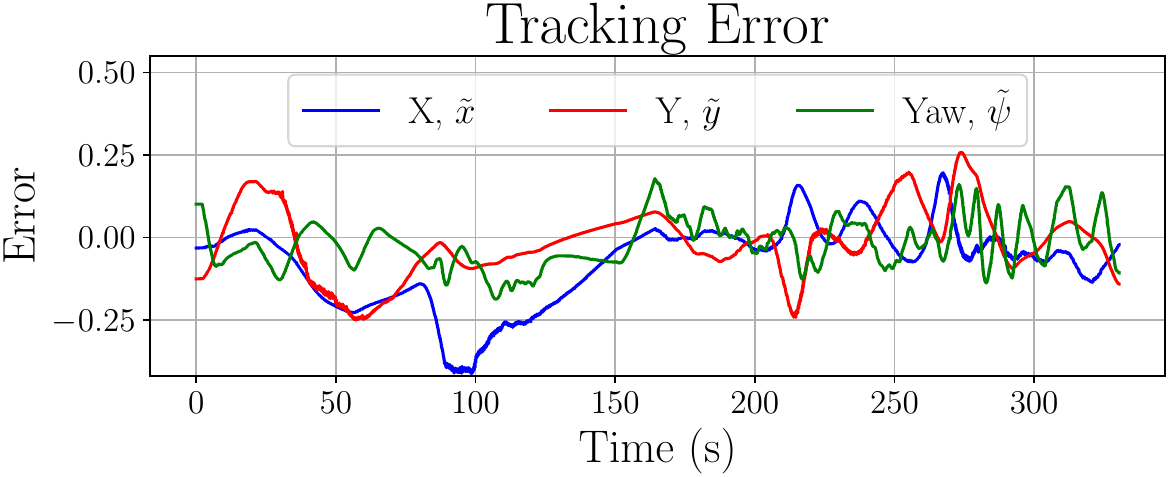}
        \caption{Basin experiment}
        \label{fig:voyager_basin_subfig}
    \end{subfigure}
    \caption{
        4-corner tests for C/S Voyager in simulation (left) and basin (right) with vessel outline plotted to scale. Trajectories show that the same control software reproduces comparable results in both environments.
    }
    \label{fig:four_corner_test_main}
\end{figure}

We conducted a series of \emph{4-corner tests} \citep{skjetne2017amos} with C/S Voyager in calm water to validate the pipeline for transferring the control architecture from simulation to the basin.
The same control software was deployed unchanged in the reduced-order simulation and on the physical model in the basin.
During the experiment, the vessel was commanded to move to the four corners of a square with a side length of $3\,\mathrm{m}$, with a different heading angle at each corner.
The resulting trajectories and tracking errors are shown in Figure~\ref{fig:four_corner_test_main}.
The position, yaw, and velocity RMSE were $0.062\,\mathrm{m}$, $0.439^\circ$, and $0.031\,\mathrm{m/s}$ in simulation, and $0.180\,\mathrm{m}$, $4.295^\circ$, and $0.036\,\mathrm{m/s}$ in the basin.
Relative to the vessel length ($L = 0.99\,\mathrm{m}$), the position RMSE is approximately $0.06\,L$ in simulation and $0.18\,L$ in the basin.
The main source of discrepancy is the unmodeled thruster dynamics and the lack of environmental disturbances in the reduced-order simulation, but the trajectories are qualitatively similar, and the control software is able to complete the task in both environments.
We consider this level of error acceptable at this stage of development, considering the purpose of model-scale testing is to verify that the algorithm works as intended on a physical vessel, and to identify areas for improvement before scaling up to a full-size vessel.
Here in this example, the vessel completed the task with the same software used in simulation, which gives confidence that the algorithm can be carried forward to a larger vessel after retuning for its dynamics and actuators.

\subsection{Autonomy verification experiments}
\begin{figure}[h]
    \centering
    \includegraphics[width=0.65\linewidth]{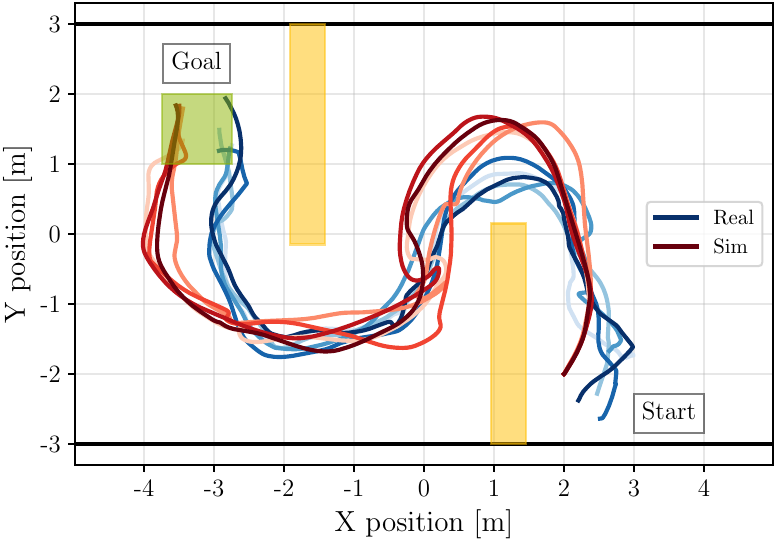}
    \caption{Example results from symbolic control for docking experiments; autonomous
        docking maneuver under symbolic control with safety guarantees \citep{dietrich2025symbolic}}
    \label{fig:formal_methods_examples}
\end{figure}

Beyond low-level GNC benchmarking, the testbed has also supported studies at the level of high-level autonomy and formal verification.
\citet{dietrich2025symbolic} used the physical testbed to validate a hierarchical symbolic control architecture for autonomous docking of a dynamic positioning vessel, combining a real-time symbolic controller for desired surge, sway, and yaw velocities with the low-level velocity feedback loop described in Section \ref{subsubsec:control_and_estimation_modules}, and demonstrated safety guarantees through physical experiments on a scaled model vessel (see Figure \ref{fig:formal_methods_examples}).
Separately, \citet{krasowski2025pacstl} used the simulation environment to run-time monitor a maritime navigation specification written in Signal Temporal Logic (STL).
Their approach extends STL with probably approximately correct (PAC) bounded reachability predictions, allowing the monitor to compute robustness bounds that account for model uncertainty.
These studies illustrate that the testbed extends beyond conventional GNC tuning to support symbolic-control and formal-verification research on the same hardware-software stack.

\section{Conclusions}
\label{sec:conclusion}

In this paper, we presented our development of a digital-physical testbed for ship autonomy research.
The contribution of the paper is (i) an open, modular software-hardware stack shared across a heterogeneous fleet of model vessels, (ii) a publicly released hydrodynamic dataset and simulation library for that fleet, and (iii) a TRL-referenced pipeline connecting model-scale testing to half-scale (R/V milliAmpere1) and full-scale (R/V Gunnerus) validation.
The infrastructure consists of a fleet of small-scale model vessels, vessel-specific simulation environments, and a digital twin implemented in Unity.
These components are tightly integrated into a modular and scalable testbed, allowing researchers to transition between simulation and physical experimentation.
The testbed enables safe, repeatable, and cost-effective testing of autonomy algorithms, while also supporting more advanced use cases such as remote control and hybrid testing scenarios.

While the digital-physical testbed offers a valuable platform for development and testing, it has several limitations.
Small-scale models are subject to Froude--Reynolds scaling conflicts, so results do not translate directly to half-scale or full-scale platforms without further study.
The higher-fidelity potential-flow model (section \ref{subsubsec:high_fidelity_simulation_model}) does not resolve viscous or strongly nonlinear effects, which is reflected in our mooring test.
Bollard-pull tests have not been conducted for the entire C/S fleet, limiting the accuracy of the thruster models.
The present experiments therefore validate the testbed workflow and software portability at model scale, rather than establishing full-scale hydrodynamic equivalence.
The pipeline does not currently propagate uncertainty from the hydrodynamic dataset through to physical experiments, and the controlled laboratory environment restricts exposure to the variability of real maritime operations.
Additionally, safety procedures and cybersecurity of remote operation were kept outside the scope of this paper, and they need to be addressed before networked operational trials targeting higher TRLs.
We plan to extend the ShipX~\citep{shipx} hydrodynamic dataset by including more vessels from the C/S fleet.
We will integrate thruster models and bollard-pull data into the simulation framework.
Additionally, there is ongoing work on developing a ship power system simulator named \emph{powersimpy}, which should be able to run in conjunction with \emph{mcsimpy} to account also for power system dynamics.

\newpage

\subsection*{Data availability statement}

The datasets and software that support the findings of this study are openly available.
The hydrodynamic datasets used for model development can be accessed at
\url{https://doi.org/10.5281/zenodo.17274088}.
The higher-fidelity potential-flow simulation framework \emph{mcsimpy} is available at
\url{https://doi.org/10.5281/zenodo.17274094},
the low-fidelity simulator \emph{shoeboxpy} at
\url{https://doi.org/10.5281/zenodo.17277253},
and the C/S software suite at
\url{https://doi.org/10.5281/zenodo.17233654}.
All repositories are publicly accessible under open-source licenses and are being continuously updated.
The cited DOIs refer to the archived releases used to produce the results in this paper.

\subsection*{Acknowledgments}

Work funded by the Research Council of Norway (RCN) through SFI AutoShip (RCN grant 309230), Norwegian Maritime AI Centre (RCN grant 359242), NTNU AMOS (RCN grant 223254), SFI Harvest (RCN grant 309661), and partly funded by the European Union via the LORELEI-X Twinning project (GA 101159489) and NTNU VISTA CAROS. The authors would also like to thank Jon Estil Krågebakk for providing 3D models, the team of engineers at MC-Lab, Robert Opland, and Vebjørn Steinsholt.

\printbibliography

\end{document}